\documentclass[letterpaper, 10 pt, conference]{ieeeconf}  

\IEEEoverridecommandlockouts                              
\usepackage{rotating}
\usepackage{array}
\usepackage{float}
\usepackage{etoolbox}

\usepackage{acronym}

\usepackage{blindtext} 

\usepackage[T1]{fontenc}

\usepackage{amsmath}
\usepackage{amssymb}

\usepackage{amsthm}
\usepackage{amsfonts}

\usepackage{graphicx}
\usepackage{textcomp}
\usepackage{setspace}
\usepackage{booktabs}
\usepackage{multirow}
\usepackage[table]{xcolor} 
\usepackage{epsfig}
\usepackage{svg}
\usepackage[ruled,noend]{algorithm2e} 
\usepackage{bm}
\usepackage{tabularx}
\usepackage{placeins} 
\usepackage{siunitx} 

\usepackage{xurl}

\newcommand{\reals}{\mathbb{R}}
\newcommand{\R}{\reals}
\newcommand{\Rnonneg}{\reals_{\geq 0}}
\newcommand{\Rplus}{\reals_{>0}}

\renewcommand{\emptyset}{\varnothing}

\newcommand{\st}{\mathrm{s.t.}}

\newcommand{\abs}[1]{\left | #1 \right |}
\newcommand{\norm}[1]{\left\Vert #1 \right \Vert}

\newcommand{\bmat}[1]{\begin{bmatrix}#1\end{bmatrix}}

\newcommand{\argmin}[1]{\underset{\substack{#1}}{\operatorname{argmin}}}

\newcommand{\Acal}{\mathcal{A}}
\newcommand{\Bcal}{\mathcal{B}}
\newcommand{\Ccal}{\mathcal{C}}

\newcommand{\Ecal}{\mathcal{E}}

\newcommand{\Gcal}{\mathcal{G}}

\newcommand{\Mcal}{\mathcal{M}}

\newcommand{\Pcal}{\mathcal{P}}

\newcommand{\Rcal}{\mathcal{R}}
\newcommand{\Scal}{\mathcal{S}}
\newcommand{\Tcal}{\mathcal{T}}
\newcommand{\Ucal}{\mathcal{U}}
\newcommand{\Vcal}{\mathcal{V}}
\newcommand{\Wcal}{\mathcal{W}}
\newcommand{\Xcal}{\mathcal{X}}
\newcommand{\Ycal}{\mathcal{Y}}
\newcommand{\Zcal}{\mathcal{Z}}

\newcommand{\bpi}{\boldsymbol{\pi}}

\newcommand{\eqn}[1]{\begin{align} #1 \end{align}}
\newcommand{\neqn}[1]{\begin{align*} #1 \end{align*}}

\newcommand{\set}[1]{\{#1\}}

\newcolumntype{g}{>{\columncolor{gray!30}}r}

\acrodef{BCH}[BCH]{Baker-Campbell-Hausdorff}
\acrodef{CBF}[CBF]{Control Barrier Function}
\acrodef{CBF-QP}[CBF-QP]{Control Barrier Function Quadratic Program}
\acrodef{CDC}[CDC]{Conference on Decision and Control}
\acrodef{CESDF}[CESDF]{Certified ESDF}
\acrodef{CLF}[CLF]{Control Lyapunov Function}
\acrodef{CVO}[C-VO]{Certified Visual Odometry}
\acrodef{DCT}[DCT]{Discrete Cosine Transform}
\acrodef{DMP}[DMP]{Distance Map Planner}
\acrodef{EKF}[EKF]{Extended Kalman Filter}
\acrodef{ESDF}[ESDF]{Euclidean Signed Distance Field}
\acrodef{EZ}[EZ]{Engagement Zone}
\acrodef{FOV}[FoV]{Field of View}
\acrodef{FPV}[FPV]{First Person View}
\acrodef{GNC}[GNC]{Graduated-Nonconvexity}
\acrodef{GP}[GP]{Gaussian Process}
\acrodef{HOCBF}[HOCBF]{Higher Order Control Barrier Function}
\acrodef{ICCBF}[ICCBF]{Input-Constrained Control Barrier Function}
\acrodef{IEEE}[IEEE]{Institute of Electrical and Electronics Engineers}
\acrodef{IMU}[IMU]{Inertial Measurement Unit}
\acrodef{ISS}[ISS]{Input-to-State}
\acrodef{KF}[KF]{Kalman Filter}
\acrodef{ML}[ML]{Machine Learning}
\acrodef{MPC}{Model Predictive Control}
\acrodef{NGPKF}[NGPKF]{Numerical Gaussian Process Kalman Filter}
\acrodef{ODE}[ODE]{Ordinary Differential Equation}
\acrodef{QP}[QP]{Quadratic Program}
\acrodef{RGBD}[RGBD]{RGB-Depth}
\acrodef{RL}[RL]{Reinforcement Learning}
\acrodef{RoS}[RoS]{Rate of Spread}
\acrodef{SDE}[SDE]{Stochastic Differential Equation}
\acrodef{SDF}[SDF]{Signed Distance Field}
\acrodef{SFC}[SFC]{Safe Flight Corridor}
\acrodef{SLAM}[SLAM]{Simultaneous Localization and Mapping}
\acrodef{SOS}[SOS]{Sum of Squares}
\acrodef{SVD}[SVD]{Singular Value Decomposition}
\acrodef{TCAC}[TCAC]{Technical Committee on Aerospace Controls}
\acrodef{TLS}[TLS]{Truncated Least Squares}
\acrodef{TSDF}[TSDF]{Truncated Signed Distance Field}
\acrodef{TSD}[TSD]{Target Spatial Distribution}
\acrodef{VIO}[VIO]{Visual Inertial Odometry}
\acrodef{VO}[VO]{Visual Odometry}
\acrodef{WLS}[WLS]{Weighted Least Squares}
\acrodef{SST}[SST]{Stable-Sparse RRT}

\theoremstyle{plain}
\newtheorem{theorem}{Theorem}
\newtheorem{corollary}{Corollary}
\newtheorem{lemma}{Lemma}

\newtheorem{definition}{Definition}

\newtheorem{assumption}{Assumption}

\theoremstyle{remark}

\AtBeginEnvironment{definition}{\let\oldemph\emph \renewcommand{\emph}[1]{\textbf{#1}}}
\AtEndEnvironment{definition}{\let\emph\oldemph}

\usepackage{hyperref}
\usepackage{cleveref}
\crefformat{problem}{Problem~#2#1#3}
\crefformat{assumption}{Assumption~#2#1#3}

\newtoggle{showproof}
\toggletrue{showproof}

\newcommand{\toggleproof}[1]{
  \iftoggle{showproof}{
  \begin{proof}
    #1
  \end{proof}
  }{
  }
}

\crefname{assumption}{assumption}{assumptions}
\Crefname{assumption}{Assumption}{Assumptions}

\crefname{definition}{Def.}{Defs.}
\Crefname{definition}{Def.}{Defs.}

\SetKwInOut{KwParam}{Parameters}

\title{\LARGE \bf Multi-Objective Kinodynamic Planning in Stochastic Environments}

\author{Thomas Marshall Vielmetti$^{1}$, Daniel Cherenson$^{1}$, and Dimitra Panagou$^{1,2}$
\thanks{This work was supported by the National Science Foundation (NSF) under grant no. 2137195.}
\thanks{$^{1}$Department of Robotics,
        University of Michigan, Ann Arbor, MI 48109, USA
      {\tt\small \{mvielmet,dmrc,dpanagou\}@umich.edu}}%
\thanks{$^{2}$Department of Aerospace Engineering,
        University of Michigan, Ann Arbor, MI 48109, USA}%
}

\begin{document}

\maketitle

\begin{abstract}
  This paper addresses kinodynamic planning for stochastic nonlinear systems subject to non-Gaussian disturbances in reactive stochastic hybrid environments.
  Conventional chance-constrained formulations require a prescribed risk threshold.
  We reformulate the problem as a multi-objective optimization in which risk and expected cost are jointly minimized.
  To efficiently approximate the Pareto front of solutions to the proposed problem, we introduce Stochastic Multi-Objective Stable Sparse RRT (SMO-SST), which extends the sparsification approach of Stable Sparse RRT (SST) to the proposed stochastic multi-objective problem. 
  We prove the proposed algorithm to be asymptotically near-optimal with respect to the risk-cost Pareto front, and provide explicit bounds on approximation error.
  In wall-clock-matched trials, the proposed algorithm outperformed non-pruning ablations, showcasing the efficacy of the proposed pruning mechanism.
  These results provide a principled framework for exploring risk-performance tradeoffs in nonlinear, non-Gaussian, reactive environments.\footnote{Code: \url{https://github.com/MarshallVielmetti/Multi_Objective_Planning_Stochastic_Environments}}
\end{abstract}

\section{Introduction}\label{sec:introduction}

This paper considers kinodynamic motion planning for discrete-time stochastic nonlinear systems operating in reactive stochastic environments.
In particular, we consider environments modeled as stochastic hybrid systems, which combine continuous state evolution with discrete modes and stochastic transitions, representing behaviors that cannot be captured naturally by continuous dynamical models alone \cite{lygerosStochasticHybridSystems2010}.
Such models arise, for example, when interacting with multimodal pedestrian behaviors \cite{parwanaRiskawareMPPIStochastic2024}, communication-constrained multi-agent systems \cite{vielmettiR3RDecentralizedMultiAgent2026}, and resource-constrained adversaries \cite{vonmollBasicEngagementZones2024}.
Planning in these settings is especially challenging because uncertainty arises not only from the dynamics of the controlled system, but also from the evolution and reaction of the surrounding environment.

Stochastic motion planning is commonly formulated as a chance-constrained optimization problem~\cite{charnesDeterministicEquivalentsOptimizing1969,schwarmChanceconstrainedModelPredictive1999}, in which expected trajectory cost is minimized subject to a prescribed upper bound on the probability of violating safety constraints.
For linear systems under Gaussian uncertainty and convex constraints, the chance constraint can be evaluated analytically, enabling efficient optimization-based approaches \cite{liProbabilisticallyConstrainedModel2002,blackmoreChanceConstrainedOptimalPath2011}.
Particle-based methods extend this framework to non-Gaussian uncertainty by approximating the state distribution through samples \cite{blackmoreProbabilisticParticleControlApproximation2010}, although efficient reformulations remain limited to restricted dynamics and constraint sets.

Sampling-based motion planners provide an alternative for problems in which directly solving the chance-constrained optimization is difficult. 
Kinodynamic RRT methods \cite{lavalleRandomizedKinodynamicPlanning1999} have been extended to stochastic systems in several ways.
LQG-MP \cite{vandenbergLQGMPOptimizedPath2011} assumes Gaussian process noise and evaluates the closed-loop behavior of candidate RRT trajectories.
CC-RRT \cite{ludersChanceConstrainedRRT2010,ludersProbabilisticFeasibilityNonlinear2011} and Monte Carlo Motion Planning (MCMP) \cite{jansonMonteCarloMotion2018} use sampled realizations to estimate probabilistic feasibility under broader uncertainty models.
More recent approaches have considered nonlinear systems and non-Gaussian disturbances \cite{lewRiskAverseTrajectoryOptimization2024}.
These methods broaden the classes of stochastic dynamics that can be considered, but generally retain a prescribed notion of acceptable risk.

Most chance-constrained formulations require an acceptable safety-risk level to be selected a priori and then optimize performance subject to that threshold.
Parallel Uncertainty-aware Multiobjective Planning (PUMP)~\cite{ichterRealtimeStochasticKinodynamic2017} explicitly recognize performance and collision probability as competing objectives, and performs multi-objective search over their Pareto front.
This avoids committing to a single cost-risk scalarization during the search process; however, the formulation considered in PUMP assumes linear dynamics with Gaussian process and measurement noise.

Thus, existing methods address subsets of the challenges considered in this work, but do not simultaneously accommodate nonlinear system dynamics, non-Gaussian uncertainty, explicit optimization of the performance-risk tradeoff, and constraints induced by reactive environments with stochastic hybrid dynamics.
This gap motivates our work, with the following contributions:

\begin{enumerate}
  \item \textbf{Problem formulation:} We formulate the planning problem as a multi-objective optimization of expected execution cost and safety-violation risk.
  \item \textbf{Algorithm and Analysis:} We introduce Stochastic Multi-Objective SST (SMO-SST), prove asymptotic near-optimality with respect to the risk-cost Pareto front, and provide an approximation bound.
  \item \textbf{Evaluation:} We evaluate SMO-SST in a pursuit-evasion scenario against stochastic adversaries.
\end{enumerate}

\iftoggle{showproof}{
\textit{Organization:} \Cref{sec:system_model} describes system models, the problem statement is presented in \Cref{sec:problem_formulation}, the method is described in \Cref{sec:methods}, analysis is conducted in \Cref{sec:analysis}, simulation results are presented in \Cref{sec:results}, and conclusions and directions for future work are given in \Cref{sec:conclusion}.
}{}

\section{System Model}\label{sec:system_model}
\emph{Notation: } Let $\R, \Rplus, \Rnonneg$ denote the sets of reals, positive reals, and non-negative reals. $\norm{\cdot}$ denotes the $\ell_2$-norm, $\abs{\cdot}$ denotes the absolute value when applied to scalars and cardinality when applied to sets.
The indicator function $\mathbf{1}_{A}(x)$ evaluates to $1$ if $x \in A$ (or condition $A$ is true) and $0$ otherwise.
$\Pcal(S)$ denotes the set of Borel probability measures on $S$, and $\Bcal(S)$ the Borel $\sigma$-algebra of $S$.

We consider kinodynamic motion planning for a stochastic nonlinear ego system operating in a reactive environment modeled as a stochastic hybrid system. 
The ego-agent is modeled by the discrete-time stochastic dynamics
\eqn{\label{eq:ego}
  x_{t+1} = f_e(x_t, u_t, w_t),
}
where $f_e : \Xcal \times \Ucal \times \Wcal \to \Xcal$ is the system dynamics, $x_t \in \Xcal \subset \R^{n_e}$ is the ego state,
$u_t \in \Ucal \subset \R^{n_u}$ is the control input,
and $w_t \in \Wcal \subset \R^{n_w}$ is the ego process disturbance.
The environment state is $\xi_t = (m_t, z_t) \in \Xi := \mathbb{M} \times \Zcal$,
where $m_t \in \mathbb{M}$ is the discrete mode
and $z_t \in \Zcal \subset \R^{n_z}$ is the continuous state.
The finite set $\mathbb{M}$ represents distinct modes of environmental behavior.
For each $m, m^\prime \in \mathbb{M}$, the environment transitions from $m$ to $m^\prime$ with probability
\eqn{\label{eq:mode_transition}
  \rho_{m,m^\prime}(x, z) := \mathbb{P}(m_{t+1} \!=\! m^\prime \!\mid\! m_t\!=\!m, x_t\!=\! x, z_t\! =\! z).
}
Conditioned on the current and next modes, the continuous environment state evolves according to
\eqn{\label{eq:environment}
  z_{t+1} = f_\xi(m_t, m_{t+1}, z_t, x_t, \nu_t)
}
where $\nu_t \in \Vcal \subset \R^{n_\nu}$ is the environment disturbance and $f_\xi : \mathbb{M} \times \mathbb{M} \times \Zcal \times \Xcal \times \Vcal \to \Zcal$ is the dynamics.
Let $\Scal:\Xi\rightrightarrows\Xcal$ be a set-valued map, where $\Scal(\xi)$ is the set of safe ego states when the environment state is $\xi\in\Xi$. Thus, a joint state $y := (x,\xi) \in \Ycal = \Xcal \times \Xi$ is safe if and only if $x\in\Scal(\xi)$, or equivalently $y\in \Ycal_\mathrm{safe}$, where
\eqn{\label{eq:joint_state_safe}
  \Ycal_\mathrm{safe} &:= \set{(x, \xi) \in \Ycal \mid x \in \Scal(\xi)}.
}

For $y = (x, m, z)$ and $y^\prime = (x^\prime, m^\prime, z^\prime)$, equip $\Ycal$ with the metric 
$d_\Ycal(y, y^\prime) := \norm{x - x^\prime} + \norm{z - z^\prime} + C_M \mathbf{1}_{m \ne m^\prime}$,
where $C_M > 0$ penalizes disagreement between discrete modes.
For $y \in \Ycal$ and $A \subset \Ycal$, let $d_\Ycal(y, A) := \inf_{y^\prime \in A} d_\Ycal(y, y^\prime)$.

Throughout, $\Xcal$ and $\Zcal$ are compact and convex, $\Ucal$, $\Wcal$, and $\Vcal$ are compact, and $\Ycal_\mathrm{safe}$ is closed.
The disturbance sequences $(w_t)_{t\geq0}$ and $(\nu_t)_{t\geq0}$ are IID, mutually independent, and independent of the initial state.
Conditional on the history through time $t$, $m_{t+1}$ is drawn from \eqref{eq:mode_transition} at the current state independently of the disturbances $w_t,\nu_t$.
For each $m,m^\prime\in\mathbb M$, define the 
one-step transition
\neqn{
  F_{m,m^\prime}(x,z,u,w,\nu)
  \!:=\!
  \bigl(
    f_e(x,u,w),
    m^\prime,
    f_\xi(m,m^\prime,z,x,\nu)
  \bigr)
  \in\Ycal.
}


\begin{assumption}[Regularity of Stochastic Dynamics]\label{assumption:dynamics_regularity}
  There exist finite constants $L_y,L_u,L_\rho,L_\partial>0$ such that:
  \begin{enumerate}
    \item For all $m,m^\prime \in \mathbb{M},\; x,x^\prime \in \Xcal, \; z,z^\prime \in \Zcal, \; u,u^\prime \in \Ucal, \; w\in\Wcal$, and $\nu \in \Vcal$,
      \neqn{
        &d_\Ycal\!\left(
          F_{m,m^\prime}(x, z, u, w, \nu),
          F_{m,m^\prime}(x^\prime, z^\prime, u^\prime, w, \nu)
          \right)\\
        &\qquad \le L_y \bigl(\norm{x - x^\prime} + \norm{z - z^\prime}\bigr) + L_u \norm{u-u^\prime}.
      }
    \item For every $m,m^\prime \in \mathbb{M}, \; x,x^\prime \in \Xcal$, and $z,z^\prime \in \Zcal$,
      \neqn{
        \abs{
          \rho_{m,m^\prime}(x, z) - \rho_{m,m^\prime}(x^\prime, z^\prime)
        } \le L_\rho \bigl(
          \norm{x-x^\prime} + \norm{z - z^\prime}
        \bigr)
      }
    \item For every $t \geq 0$, $m,m^\prime \in \mathbb{M}, \; x \in \Xcal, z \in \Zcal, u \in \Ucal$, and $\epsilon \geq 0$ satisfying $\rho_{m,m^\prime}(x, z) > 0$,
      \neqn{
        \mathbb{P}\!\left(
          d_\Ycal \left(F_{m,m^\prime}(x, z, u, w_t, \nu_t), \partial \Ycal_\mathrm{safe} \right) \leq \epsilon
          \right) 
          \leq \min\set{1, L_\partial \epsilon}.
      }
  \end{enumerate}
\end{assumption}


Define $D_\Ycal := \max\set{1,\sup_{y,y^\prime \in \Ycal} d_\Ycal(y, y^\prime)}$, so $0 < D_\Ycal < \infty$ and $d_\Ycal(y, y^\prime) \le D_\Ycal$ for all $y,y^\prime \in \Ycal$.
Finally, define the augmented joint state space $\Ycal_b := \set{(y, b) \in \Ycal \times \set{0, 1}}$, where $b$ is a safety-failure indicator, evolving by
\eqn{
  b_0 := \mathbf{1}_{y_0 \notin \Ycal_\mathrm{safe}}, \quad b_{t+1} := b_t \lor \mathbf{1}_{y_{t+1} \notin \Ycal_\mathrm{safe}},
  \label{eq:safety_indicator}
}
and the metric $d_b((y, b), (y^\prime, b^\prime)) := \frac{d_\Ycal(y, y^\prime)}{D_\Ycal} + \abs{b - b^\prime}$.
\subsection{Feedback Policy Sequences}

We formulate the planning problem over finite sequences of feedback policies and execution durations.  
We assume that the joint state $y_t$ is available to the ego agent when selecting its control input. Let $\Pi$ denote a space of admissible feedback policies $\pi : \Ycal \to \Ucal$.
In general, we do not consider planning over an arbitrary set of functions, but a set of policies parameterized by some $\theta \in \Theta \subset \R^{n_\theta}$, such that $\Pi = \set{\pi_\theta : \theta \in \Theta}$.

This approach has been considered by kinodynamic planning methods for both stabalizing unstable nonlinear systems and improving sample efficiency~\cite{perezLQRRRTOptimalSamplingbased2012}. 
It has also been explored in stochastic environments to counteract the effects of process noise on open-loop trajectory execution \cite{vandenbergLQGMPOptimizedPath2011,jansonMonteCarloMotion2018}.
Here, in addition to counteracting the effects of process noise, we are motivated by the ability of a feedback policy to react to the evolving environment.
\begin{definition}[Policy Sequence]
  Let $\tau_{\max}\in\mathbb{N}_{>0}$.
  A \emph{policy sequence} of length $k:=k(\sigma)\in\mathbb{N}_{\geq0}$ is
  \eqn{
    \sigma:=\bigl((\pi_{i},\tau_{i})\bigr)_{i=1}^k
    \in\left(\Pi\times\set{1,\ldots,\tau_{\max}}\right)^k.
  }
  Let $T_0:=0$, let $T_j:=\sum_{i=1}^j\tau_{i}$ for $1\leq j\leq k$, and set $T(\sigma):=T_k$.
  For each $0\leq t<T(\sigma)$, 
  \eqn{
    k_t&:=\min\set{j\in\set{1,\ldots,k}:t<T_j},\notag\\
    u_t&:=\pi_{k_t}(y_t).\label{eq:control_law}
  }
\end{definition}

Define $\bpi:=(\pi_{i})_{i=1}^k$ and $\Tcal:=(\tau_{i})_{i=1}^k$.
Equip $\Pi$ with the uniform metric
$d_\Pi(\pi,\pi^\prime):=\sup_{y\in\Ycal}\norm{\pi(y)-\pi^\prime(y)}$.
For $\bpi,\bpi^\prime \in\Pi^k$, 
\eqn{
  D(\bpi,\bpi^\prime):=
  \begin{cases}
    0, & k=0,\\
    \displaystyle\max_{i\in\set{1,\ldots,k}}
      d_\Pi(\pi_{i},\pi^\prime_{i}), & k\geq1.
  \end{cases}
}
\begin{assumption}[Policy Space]\label{assumption:policy_space}
  The policy space $(\Pi, d_\Pi)$ is compact, and $\exists L_\pi > 0$ such that $\forall \pi \in \Pi$, and $\forall y, y^\prime \in \Ycal$ with $m=m^\prime$,
  \neqn{
    \norm{\pi(y) - \pi(y^\prime)} \le L_\pi d_\Ycal(y,y^\prime).
  }
  Finally, let $\mu_\pi\in\Pcal(\Pi)$ satisfy
  \neqn{
    \forall \pi\in\Pi,\ \forall h>0,\qquad
    \mu_\pi\!\left(
      \set{\pi^\prime\in\Pi:
        d_\Pi(\pi^\prime,\pi)<h}
    \right)>0,
  }
  i.e. $\mu_\pi$ has full support.
\end{assumption}

Define $\Sigma(n) := \bigcup_{k=0}^n (\Pi \times \set{1, \dots, \tau_\mathrm{max}})^k$ to be the set of policy sequences of length at most $n$. 
Furthermore, for $\sigma, \sigma^\prime$, let $(\sigma \Vert \sigma^\prime) = ((\pi_1, \tau_1), \dots, (\pi_{k(\sigma)}, \tau_{k(\sigma)}), (\pi_1^\prime, \tau_1^\prime), \dots, (\pi^\prime_{k(\sigma^\prime)}, \tau^\prime_{k(\sigma^\prime)}))$.

\section{Problem Formulation}\label{sec:problem_formulation}
Let $(x_\mathrm{init}, \xi_\mathrm{init}) = y_\mathrm{init} \in \Ycal$ denote the known initial joint state.
Fix a maximum policy-sequence length $\bar{K} \in \mathbb{N}_{>0}$ and define the solution space
$\Sigma := \Sigma(\bar{K})$.

Executing a policy sequence $\sigma\in\Sigma$ from $y_0=y_\mathrm{init}$ under
\eqref{eq:ego}, \eqref{eq:mode_transition}, \eqref{eq:environment},
\eqref{eq:control_law}, and \eqref{eq:safety_indicator}
generates the random augmented-state trajectory
$\bigl((y_t,b_t)\bigr)_{t=0}^{T(\sigma)}$.
Its probability law
$\mathbb{P}_{y_\mathrm{init}}^\sigma\in\Pcal(\Ycal_b^{T(\sigma)+1})$
assigns to each measurable set of trajectories
$A\in\Bcal(\Ycal_b^{T(\sigma)+1})$ the probability
\eqn{\label{eq:probability_law}
  \mathbb{P}_{y_\mathrm{init}}^\sigma(A)
  := \mathbb{P}\!\left(
    \bigl((y_t,b_t)\bigr)_{t=0}^{T(\sigma)}\in A
  \right),
}
where $\mathbb{P}$ governs the disturbances and stochastic mode transitions.
We write $\mathbb{E}_{y_\mathrm{init}}^\sigma$ for expectation with respect to this law.

We evaluate each policy sequence by its probability of safety-constraint violation and expected execution cost.

\begin{definition}[Risk-Cost Objective]\label{def:multi_objective_cost_mapping}
  Let $T=T(\sigma)$. Let $J : \Sigma \to [0, 1] \times \Rnonneg$ be the vector-valued objective
  \eqn{\hspace{-0.35cm}
    J(\sigma) \!:=\!
      \bmat{
        J_\mathrm{risk}(\sigma)\\
        J_\mathrm{cost}(\sigma)
      } \!\!=\!\! \bmat{
        \mathbb{E}_{y_\mathrm{init}}^\sigma [b_T]\\
        \mathbb{E}_{y_\mathrm{init}}^\sigma\! \left[
          \sum_{t=0}^{T-1}\ell(x_t, u_t) \!+\! \ell_T(x_T)\!
        \right]\!
      }\!\!,
  }
  where $\ell : \Xcal \times \Ucal \to \Rnonneg$ is the running cost and $\ell_T : \Xcal \to \Rnonneg$ is the terminal cost.
  For $\sigma=\varnothing$, the horizon is zero and the running-cost sum is empty, so
  $J(\varnothing)=\bmat{b_0 & \ell_T(x_\mathrm{init})}^{\top}$.
\end{definition}
The first objective is the probability of violating the safety constraint, while the second is the expected execution cost.
We treat these as competing objectives, so to compare candidate solutions, we use the notion of Pareto dominance.

\begin{assumption}[Cost Continuity]\label{assumption:cost_regularity}
  There exist finite constants $L_\ell, L_{\ell_T} > 0$ such that $\forall x, x^\prime \in \Xcal, \; u,u^\prime \in \Ucal$,
  \neqn{
    \abs{\ell(x, u) - \ell(x^\prime, u^\prime)} &\le L_\ell \left(\norm{x - x^\prime} + \norm{u - u^\prime}\right)\\
    \abs{\ell_T(x) - \ell_T(x^\prime)} &\le L_{\ell_T} \norm{x - x^\prime}.
  }
\end{assumption}

\begin{definition}[Pareto Dominance]\label{def:pareto_dominance}
  Let $\sigma, \sigma^\prime \in \Sigma$. Then $\sigma$ \emph{dominates} $\sigma^\prime$, denoted $\sigma \prec \sigma^\prime$, if
  \neqn{
    J_\mathrm{risk}(\sigma) \leq J_\mathrm{risk}(\sigma^\prime), \quad
    J_\mathrm{cost}(\sigma) \leq J_\mathrm{cost}(\sigma^\prime),
  }
  with at least one inequality holding strictly.
\end{definition}

\begin{definition}[Pareto Optimal Set \& Pareto Front]\label{def:pareto_optimal}
  The Pareto-optimal set is 
  \eqn{
    \Sigma^* := \set{\sigma \in \Sigma \mid \nexists \sigma^\prime \in \Sigma \text{ s.t. } \sigma^\prime \prec \sigma},
  }
  and $J(\Sigma^*)$ is the corresponding \emph{Pareto front}.
\end{definition}

The stochastic multi-objective planning problem is
\begin{equation}\label{eq:multi_objective_optimization}
\begin{aligned}
  \min_{\sigma \in \Sigma} \quad J(\sigma) \; \st \; y_0 = y_\mathrm{init},
\end{aligned}
\end{equation}
where $\min$ is interpreted in the Pareto sense.
Our objective is to construct a finite set of policy sequences whose objective vectors approximate the Pareto front. 
\begin{definition}[Asymptotic Near-Pareto Optimality]
\label{def:asymptotic_near_optimality}
  Let a tolerance vector $\boldsymbol{\epsilon}:=\bmat{\epsilon_q & \epsilon_c}^{\top} \in \Rnonneg^2$.
  Let $\Sigma^{(n)}\subseteq\Sigma$ denote the finite set of candidate policy sequences retained by the algorithm after iteration $n$.
  The algorithm is \emph{asymptotically
  near-Pareto-optimal} if $\forall \eta_q,\eta_c>0$, there exists an iteration
  $n_{\eta_q,\eta_c}<\infty$ such that
  $\forall n\ge n_{\eta_q,\eta_c}, \; \forall \sigma^* \in \Sigma^*$, there exists
  $\sigma_n\in\Sigma^{(n)}$ satisfying
  \eqn{\label{eq:asymptotically_near_optimal}
    J_\mathrm{risk}(\sigma_n)
    &\le
    J_\mathrm{risk}(\sigma^*)+\epsilon_q+\eta_q,\\
    J_\mathrm{cost}(\sigma_n)
    &\le
    J_\mathrm{cost}(\sigma^*)+\epsilon_c+\eta_c.
  }
\end{definition}



\section{Methods}\label{sec:methods}
\begin{figure*}[ht]
  \centering
  \includegraphics[width=0.99\textwidth]{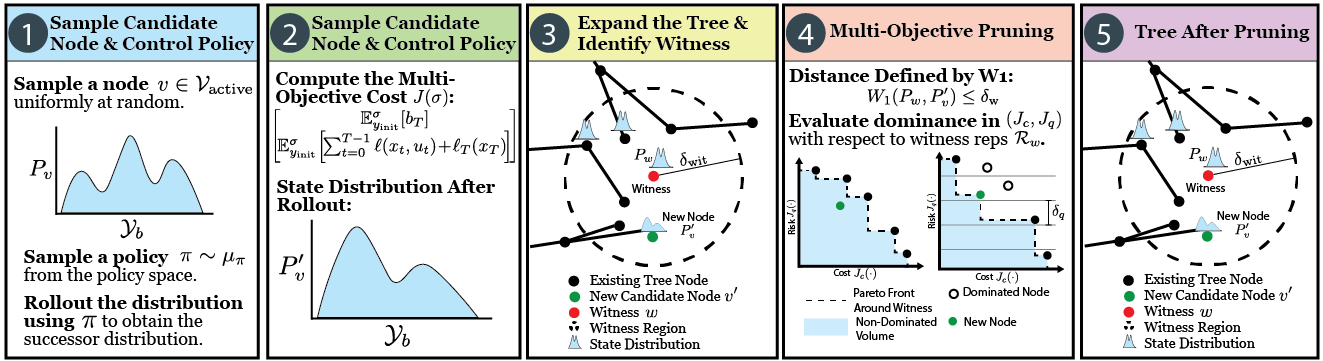}
  \vspace{-0.25em}
  \caption{One iteration of SMO-SST. From left to right, the planner samples and propagates a node-policy pair, assigns the resulting candidate to a witness using $W_1$ distance, and applies multi-objective pruning to update the witness representatives and search tree.}
  \label{fig:algo}
\end{figure*}

We introduce SMO-SST, a stochastic, multi-objective extension of \ac{SST}~\cite{liSparseMethodsEfficient2015}. An overview of the proposed method is given in \cref{fig:algo}.
We first briefly introduce \ac{SST}, redefine the core datatypes and structures for our multi-objective environment, introduce our approach to pruning in our stochastic multi-objective environment, and finally introduce the proposed algorithm and pseudocode.

\subsection{Preliminaries: \ac{SST}}
\ac{SST}~\cite{liSparseMethodsEfficient2015} is an asymptotically near-optimal sampling-based motion planner that approximately solves the motion planning problem for nonlinear systems.
\ac{SST} incrementally constructs a directed search tree $\Gcal = (\Vcal, \Ecal)$.
Each node represents a reachable state together with the accumulated cost of the control trajectory to reach it, and each edge represents a dynamically feasible trajectory segment generated by propagating a fixed control input.
\ac{SST} uses witnesses to maintain a sparse approximation of the reachable state space.
Each witness defines a neighborhood of radius $\delta_\mathrm{w}$ and maintains a single representative node.
A newly generated node replaces the representative when it lies within the witness neighborhood and has lower accumulated cost, making the replaced node inactive and ineligible for expansion.
Nodes that are no longer active are deleted once they are not required as ancestors of retained trajectories. 

\subsection{Extension to Stochastic Environments}
We adapt the SST framework to solve the stochastic multi-objective planning problem in \eqref{eq:multi_objective_optimization}.
We begin by redefining nodes and witnesses.

\begin{definition}[Node]\label{def:node}
  A \emph{node} $v=(k_v, t_v, P_v, J_v, \sigma_v)$ is associated with a policy sequence $\sigma_v$.
  Its depth and execution horizon are $k_v := k(\sigma_v)$ and $t_v=T(\sigma_v)$, respectively.
  Executing $\sigma_v$ from $y_\mathrm{init}$ induces the law 
  $P_v(A):= \mathbb{P}_{y_\mathrm{init}}^{\sigma_v}\bigl((y_{t_v},b_{t_v})\in A\bigr),\; A\in\Bcal(\Ycal_b)$, where $\mathbb{P}_{y_\mathrm{init}}^{\sigma_v}$ is as defined in \eqref{eq:probability_law}.
The node also stores the objective vector $J_v := \bmat{q_v & c_v}^\top = J(\sigma_v)$ as defined by \cref{def:multi_objective_cost_mapping}.
\end{definition}

The empty sequence generates the depth-zero root
\eqn{
  v_0:=\left(0,0,\delta_{(y_\mathrm{init},b_0)},J(\varnothing),\varnothing\right),
}
which is never pruned and remains available for expansion. 
$\Vcal^{(n)}$ denotes the set of nodes at iteration $n$.
\begin{definition}[Witness]\label{def:witness}
  A \emph{witness} $w=(P_w,k_w,\Rcal_w)$ consists of a fixed center distribution $P_w \in \Pcal(\Ycal_b)$, depth $k_w$, and set of representative nodes $\Rcal_w$. Only nodes with depth $k_v = k_w$ may belong to $\Rcal_w$.
\end{definition}
$\Wcal_\mathrm{wit}^{(n)}$ denotes the witness set at iteration $n$.

Witnesses partition the search space into balls of radius $\delta_\mathrm{w}$.
In \ac{SST}, the ball is a Euclidean ball in the state space, i.e. distance is defined by the usual metric in $\R^n$; we begin by extending this notion to $\Pcal(\Ycal_b)$.
\begin{definition}[Wasserstein Distance]\label{def:wasserstein}
  For $P,P'\in\Pcal(\Ycal_b)$, define the $1$-Wasserstein distance
  \eqn{\label{eq:wasserstein}
    W_1(P,P'):=\inf_{\lambda\in\Gamma(P,P')}
    \int_{\Ycal_b\times\Ycal_b}d_b(s,s')\,d\lambda(s,s'),
  }
  where $\Gamma(P, P^\prime)$ is the set of couplings with marginals $P$ and $P'$.
\end{definition}

Let $v$ be a newly generated node at iteration $n$.
If $\exists w \in \Wcal_\mathrm{wit}^{(n)}, k_v = k_w$ such that $W_1(P_v, P_w) \le \delta_\mathrm{w}$, then $v$ is considered for insertion into $\Rcal_w$. Otherwise, a new witness node is created, centered at $P_v$.

The purpose of the representative set is to maintain sparsity of the search graph.
In \ac{SST}, $\Rcal_w$ maintains only a single representative, specifically the node with the lowest cost that has been considered for insertion.
We now introduce a Pareto update law, \texttt{SparseParetoUpdate}, that considers a multi-objective cost but still ensures finitely many nodes are maintained by each witness.
Let $v$ be a newly generated node at iteration $n$ assigned to witness $w \in \Wcal_\mathrm{wit}^{(n)}$ such that $k_v= k_w, \; \text{ and } \; W_1(P_v, P_w) \le \delta_\mathrm{w}$.
Let $\bar{\Rcal}_w = \Rcal_w \cup \set{v}$. We first compute the Pareto-set of $\bar{\Rcal}_w$, i.e.
\eqn{\label{eq:pareto_representative_update}
  \hat{\Rcal}_w = \set{v \in \bar{\Rcal}_w \mid \not\exists s \in \bar{\Rcal}_w, v \ne s, \sigma_s \prec \sigma_v }.
}
However, the Pareto-optimal set may contain infinitely many solutions.
One common approach to sparsify the Pareto set is the notion of $\epsilon$-dominance, which introduces an additive sparsification parameter to \cref{def:pareto_dominance}, and is the approach adopted in~\cite{razzaqMultiObjectiveKinodynamicMotion2026}. 
For a resolution $\delta_q\in(0,1]$, define $g(q):=\left\lfloor q/\delta_q\right\rfloor, \; N_q:=\left\lfloor1/\delta_q\right\rfloor+1$.
For each $i\in\set{0,\ldots,N_q-1}$, let $\hat{\Rcal}_w^{(i)}:=\set{v\in\hat{\Rcal}_w\mid g(q_v)=i}$.
In each occupied cell, retain exactly one representative $v_i\in\arg\min_{v\in\hat{\Rcal}_w^{(i)}}\;c_v$, breaking ties first by lower risk and then by earlier node creation. The resulting set $\Rcal_w^* := \bigcup_{i=0}^{N_q-1} \set{v_i}$ is a subset of \eqref{eq:pareto_representative_update} and thus mutually non-dominated, and satisfies $\abs{\Rcal_w^*} \le N_q$.

Any nodes $v \in \Rcal_w \setminus \Rcal_w^*$ are \texttt{Prune}d.
They are first marked as \texttt{inactive}, then if the node is a leaf, it is recursively deleted until an active or non-leaf ancestor is reached.
 
\subsection{Stochastic Multi-Objective SST (SMO-SST)}
The full pseudocode is given in \cref{alg:pop_smo_sst}.
SMO-SST initializes a search tree and depth-zero witness at the initial state.
At each iteration, it uniformly selects an active node $v \in \Vcal_\mathrm{active}$, samples a feedback policy $\pi \sim \mu_\pi$ and execution duration $\tau \sim \operatorname{Uniform}(\set{1, \dots, \tau_\mathrm{max}})$, then propagates the $P_v$ through the closed-loop dynamics in \eqref{eq:control_law}.
The operation $v^\prime\gets\texttt{Propagate}(v,\pi,\tau)$ constructs the child node with $k_{v^\prime}=k_v+1$, $t_{v^\prime}=t_v+\tau$, and $\sigma_{v^\prime}=\sigma_v\Vert(\pi,\tau)$.
It computes $P_{v^\prime}$, $q_{v^\prime}=\int b\,dP_{v^\prime}$, and
\eqn{\label{eq:terminal_cost_update}
  c_{v^\prime}=c_v+\mathbb E\!\left[
    \sum_{t=t_v}^{t_{v^\prime}-1}\ell(x_t,u_t)
    + \ell_T(x_{t_{v^\prime}})-\ell_T(x_{t_v})\right].
}
The expectation is under execution of $\sigma_{v^\prime}$.
The previous terminal contribution is replaced, thus $c_v$ may decrease along an edge.
The planner returns the graph of active nodes and their surviving inactive ancestors.

\begin{algorithm}[t]
\caption{SMO-SST}
\label{alg:pop_smo_sst}
\footnotesize
\KwParam{$\bar{K},\tau_{\max},\delta_{\mathrm{w}},\delta_q,N_{\mathrm{iterations}}$}
\KwIn{$y_{\mathrm{init}},\mu_\pi$}
$v_0\gets\texttt{Node}(0,0,\delta_{(y_{\mathrm{init}},b_0)},
  J(\varnothing),\varnothing)$\;
$\Wcal_{\mathrm{wit}}\gets\set{\texttt{Witness}(P_{v_0},0,\set{v_0})}$\;
$\Vcal_{\mathrm{active}}\gets\set{v_0},\quad
  \Vcal_{\mathrm{inactive}}\gets\emptyset,\quad\Ecal\gets\emptyset$\;
\For{$n=1$ \KwTo $N_{\mathrm{iterations}}$}{
  $v\sim\mathrm{Uniform}(\set{v\in\Vcal_{\mathrm{active}}\mid k_v<\bar{K}})$\;
  $\pi\sim\mu_\pi,\quad\tau\sim\mathrm{Uniform}(\set{1,\ldots,\tau_{\max}})$\;
  $v'\gets\texttt{Propagate}(v,\pi,\tau)$\;
  $\Wcal_{\mathrm{depth}}\gets\set{w\in\Wcal_{\mathrm{wit}}\mid k_w=k_{v'}}$\;
  \eIf{$\Wcal_{\mathrm{depth}}=\emptyset$}{
    $w\gets\texttt{Witness}(P_{v'},k_{v'},\emptyset)$\;
    $\Wcal_{\mathrm{wit}}\gets\Wcal_{\mathrm{wit}}\cup\set{w}$\;
  }{
    $w\gets\argmin{w'\in\Wcal_{\mathrm{depth}}}W_1(P_{v'},P_{w'})$\;
    \If{$W_1(P_{v'},P_w)>\delta_{\mathrm{w}}$}{
      $w\gets\texttt{Witness}(P_{v'},k_{v'},\emptyset)$\;
      $\Wcal_{\mathrm{wit}}\gets\Wcal_{\mathrm{wit}}\cup\set{w}$\;
    }
  }
  $\Rcal_{\mathrm{old}}\gets\Rcal_w$\;
  $\Rcal_w\gets\texttt{SparseParetoUpdate}(\Rcal_w,v',\delta_q)$\;
  \If{$v'\in\Rcal_w$}{
    $\Vcal_{\mathrm{active}}\gets\Vcal_{\mathrm{active}}\cup\set{v'}$\;
    $\Ecal\gets\Ecal\cup\set{(v,v')}$\;
  }
  \ForEach{$r\in\Rcal_{\mathrm{old}}\setminus\Rcal_w$}{
    $\texttt{Prune}(\Vcal_{\mathrm{active}},\Vcal_{\mathrm{inactive}},\Ecal,r)$\;
  }
}
\Return{$\Gcal=(\Vcal_{\mathrm{active}}\cup\Vcal_{\mathrm{inactive}},\Ecal)$}\;
\end{algorithm}

\section{Analysis}\label{sec:analysis}
We analyze the theoretical properties of the proposed algorithm in three stages, with the goal of proving it satisfies asymptotic near-Pareto-optimality as in \cref{def:asymptotic_near_optimality}.
We first prove Lipschitz continuity of the multi-objective mapping $J(\sigma)$, then show that pruning retains quality nodes while also maintaining sparsity, which together let us show the desired property.
All results take \cref{assumption:dynamics_regularity,assumption:policy_space,assumption:cost_regularity} to be true.

\subsection{Continuity of the Objective Mapping}
We begin by controlling trajectory divergence under nearby policy sequences.
The first step conditions on identical environment-mode transitions.
\begin{definition}[Discrete Mode Sequence]\label{def:mode_sequence}
  For a finite horizon $T$, let $M = (m_0, \dots, m_T) \in \Mcal_T \subset \mathbb{M}^{T+1}$ denote an admissible environment mode sequence.
\end{definition}
\begin{lemma}[Bounded Conditional Trajectory Divergence]\label{lemma:bounded_trajectory_divergence}
  Let a duration sequence $\Tcal$ be of length $k$ with horizon $T$, a mode sequence $M\in\Mcal_T$, and disturbance sequences $(w_t)_{t\geq0}$ and $(\nu_t)_{t\geq0}$.
  Let $y_t,y_t^\prime$ be the trajectories generated from initial conditions $y_0,y_0^\prime\in\Ycal$, with $m_0=m_0^\prime$, by policy sequences $\bpi,\bpi^\prime \in \Pi^k$.
  If
  \neqn{
    d_\Ycal(y_0, y_0^\prime) + D(\bpi, \bpi^\prime) \le \delta,
  }
  then $d_\Ycal(y_t,y_t^\prime)\le C_t\delta$ for every $t\le T$,
  where $C_t:=(L_y + L_u L_\pi)^t + L_u \sum_{j=0}^{t-1} (L_y + L_u L_\pi)^j < \infty$.
\end{lemma}
\toggleproof{
  If $k=0$, then $T=0$ and the conclusion follows from the initial bound with $C_0=1$.
  Hence, suppose $k\geq1$.
  Because the trajectories are conditioned on the same mode sequence, their discrete modes agree at every time and contribute no mode-mismatch penalty to $d_\Ycal$.
  Let $k_t$ denote the active policy index.
  \Cref{assumption:dynamics_regularity} gives
  \neqn{
    d_\Ycal(y_{t+1}, y_{t+1}^\prime) \!\le\! L_y d_\Ycal(y_t, y_t^\prime) \!+\! L_u \norm{\pi_{k_t}(y_{t})\!-\! \pi_{k_t}^\prime(y_{t}^\prime)}.
  }
  Applying \cref{assumption:policy_space} yields
  \neqn{
    \norm{\pi_{k_t}(y_t) - \pi_{k_t}^\prime(y_t^\prime)} \le L_\pi d_\Ycal(y_t, y_t^\prime) + D(\bpi, \bpi^\prime).
  }
  Therefore,
  \neqn{
    d_\Ycal(y_{t+1}, y_{t+1}^\prime) \le (L_y + L_u L_\pi) d_\Ycal(y_t, y_t^\prime) + L_u D(\bpi, \bpi^\prime).
  }
  Let $a := L_y + L_u L_\pi$. Unrolling this recurrence yields
  \neqn{
    d_\Ycal(y_t, y_t^\prime) \le a^t d_\Ycal(y_0, y_0^\prime) + L_u \sum_{j=0}^{t-1} a^j D(\bpi, \bpi^\prime).
  }
  Taking $C_t:=a^t + L_u \sum_{j=0}^{t-1} a^j$, with $C_0 = 1$, completes the proof.
}

The conditional trajectory bound applies while the two mode sequences agree.
The next lemma shows that initial states and policies admit a coupling under which the probability of disagreement is bounded proportionally to the distance between the initial states and between the policies.
\begin{lemma}[Mode Sequence Coupling]\label{lemma:mode_probabilities}
  Fix a $\Tcal$ of length $k\leq \bar{K}$ with horizon $T$, policies $\bpi,\bpi^\prime\in\Pi^k$, and initial states $y_0,y_0^\prime\in\Ycal$ with $m_0=m_0^\prime$.
  Let $M,M^\prime\in\Mcal_T$ denote the resulting mode sequences, and define
  $K_M :=
    \frac{\abs{\mathbb M}L_\rho}{2}
    \sum_{t=0}^{\bar K\tau_{\max}-1}C_t
    <\infty$.
  There is a coupling of the systems such that
  \neqn{
    d_\Ycal(y_0,y_0^\prime)+D(\bpi,\bpi^\prime)\leq\delta
    \implies \mathbb{P}(M\ne M^\prime)\leq K_M\delta
  }
  uniformly over initial states, policies, and horizons.
\end{lemma}
\toggleproof{
  If $k=0$, then $T=0$, and $M=M^\prime$ because $m_0=m_0^\prime$.
  Hence, suppose $k\geq1$.

  Couple the two systems using the same continuous disturbances, and let $\mathcal H_t$ denote their history through time $t$.
  For each $t\in\set{0,\ldots,T-1}$, define the event the mode sequences have agreed up to time $t$ as
  \neqn{
    E_t:=\set{m_s=m_s^\prime,\;s=0,\ldots,t}.
  }
  Define the next-mode probability mass functions
  \neqn{
    p_t(\bar m)&:=\rho_{m_t,\bar m}(x_t,z_t),\notag\\
    p_t^\prime(\bar m)&:=\rho_{m_t^\prime,\bar m}(x_t^\prime,z_t^\prime),
    \qquad \bar m\in\mathbb M.
  }
  On $E_t$, draw the next modes $(m_{t+1},m_{t+1}^\prime)$ from a maximal coupling of $p_t$ and $p_t^\prime$.
  and define the event that they disagree
  \neqn{
    D_{t+1}
    :=
    E_t\cap\set{m_{t+1}\neq m_{t+1}^\prime}.
  }
  Thus, $D_{t+1}$ is the event that the first mode disagreement occurs at time $t+1$.

  On $E_t$, the systems have experienced the same mode sequence and the same continuous disturbances through time $t$.
  Therefore, \cref{lemma:bounded_trajectory_divergence} gives
  \neqn{
    d_\Ycal(y_t,y_t^\prime)\leq C_t\delta.
  }
  Define the total variation distance between probability mass functions $p$ and $q$ on $\mathbb M$ by
  \neqn{
    d_{\mathrm{TV}}(p,q)
    :=
    \frac{1}{2}\sum_{\bar m\in\mathbb M}
    \abs{p(\bar m)-q(\bar m)}.
  }
  Because the current modes agree on $E_t$, the $L_\rho$-Lipschitz continuity of the mode-transition probabilities implies
  \begin{align}
    d_{\mathrm{TV}}(p_t,p_t^\prime)
    &=
    \frac{1}{2}\sum_{\bar m\in\mathbb M}
    \abs{
      \rho_{m_t,\bar m}(x_t,z_t)
      -
      \rho_{m_t^\prime,\bar m}(x_t^\prime,z_t^\prime)
    }\notag\\
    &\leq \frac{1}{2} \sum_{\bar{m} \in \mathbb M} L_\rho d_\Ycal(y_t, y^\prime_t)\\
    &\leq
    \frac{\abs{\mathbb M}L_\rho}{2}
    d_\Ycal(y_t,y_t^\prime)\notag\\
    &\leq
    \frac{\abs{\mathbb M}L_\rho}{2}C_t\delta.
  \end{align}
  By the defining property of maximal coupling, on $E_t$,
  \neqn{
    \mathbb P\!\left(
      m_{t+1}\neq m_{t+1}^\prime
      \mid\mathcal H_t
    \right)
    =
    d_{\mathrm{TV}}(p_t,p_t^\prime).
  }
  The tower property gives
  \begin{align}
    \mathbb P(D_{t+1})
    &=
    \mathbb E\!\left[
      \mathbf 1_{E_t}
      \mathbb P\!\left(
        m_{t+1}\neq m_{t+1}^\prime
        \mid\mathcal H_t
      \right)
    \right]\notag\\
    &=
    \mathbb E\!\left[
      \mathbf 1_{E_t}
      d_{\mathrm{TV}}(p_t,p_t^\prime)
    \right]\notag\\
    &\leq
    \frac{\abs{\mathbb M}L_\rho}{2}C_t\delta
    \mathbb P(E_t)\notag\\
    &\leq
    \frac{\abs{\mathbb M}L_\rho}{2}C_t\delta.
  \end{align}

  Because $m_0=m_0^\prime$, the events $D_1,\ldots,D_T$ are disjoint, and their union is $\set{M\neq M^\prime}$.
  Therefore,
  \neqn{
    \mathbb P(M\neq M^\prime)
    =
    \sum_{t=0}^{T-1}\mathbb P(D_{t+1})
    \leq
    \frac{\abs{\mathbb M}L_\rho}{2}
    \left(\sum_{t=0}^{T-1}C_t\right)\delta.
  }
  Since $T\leq\bar K\tau_{\max}$,
  \neqn{
    \mathbb P(M\neq M^\prime)\leq K_M\delta. && \qedhere
  }
}

This result shows that for two initial states executing two policy sequences, the probability of their modes diverging is bounded by a constant multiple of the sum of the distance between the initial states and the policy sequences. 
If we replace deterministic initial states with initial state distributions, we can use this result to bound the $W_1$ distance between the distributions after executing distinct policy sequences.

\begin{theorem}[Lipschitz Continuity in Wasserstein Distance]\label{thm:wasserstein_sensitivity}
  Let $u,v$ be nodes with $k_u=k_v$.
  Let $\sigma=((\pi_i,\tau_i))_{i=1}^r,\; \sigma^\prime=((\pi_i^\prime,\tau_i))_{i=1}^r$,
  be admissible continuations of the nodes' policy sequences, where $0\le r\le\bar K-k_u$, and write $\bpi=(\pi_i)_{i=1}^r$ and $\bpi^\prime=(\pi_i^\prime)_{i=1}^r$.
  Define $\Delta:=W_1(P_u,P_v)+D(\bpi,\bpi^\prime)$.
  There exist constants $L_{\mathrm{risk}},L_{\mathrm{cost}}>0$,
  uniform over all such nodes and continuations, such that
  \eqn{
    \abs{
      J_{\mathrm{risk}}(\sigma_u\Vert\sigma)
      -J_{\mathrm{risk}}(\sigma_v\Vert\sigma^\prime)
    }
    &\!\le\! L_{\mathrm{risk}}\Delta,
    \label{eq:law_policy_sensitivity}\\
    \abs{
      \left(J_{\mathrm{cost}}(\sigma_u\Vert\sigma)-c_u\right)
      \!-\!\left(J_{\mathrm{cost}}(\sigma_v\Vert\sigma^\prime)-c_v\right)
    }
    &\!\le\! L_{\mathrm{cost}}\Delta.
    \label{eq:cost_increment_sensitivity}
  }
\end{theorem}
\toggleproof{
  Fix any $\lambda\in\Gamma(P_u,P_v)$ and, for
  every coupled pair $(y,b),(y^\prime,b^\prime)$, execute the two
  continuations from these states.
  Let $T:=\sum_{i=1}^r\tau_i$, let $y_t=(x_t,m_t,z_t)$ and
  $y_t^\prime=(x_t^\prime,m_t^\prime,z_t^\prime)$ denote the resulting
  coupled trajectories, and set $b_0=b$ and $b_0^\prime=b^\prime$.
  Define $E_T:=\set{m_t=m_t^\prime,\;t=0,\ldots,T}$.

  If the initial modes agree, put
  $\delta:=d_\Ycal(y,y^\prime)+D(\bpi,\bpi^\prime)$.
  By \cref{lemma:bounded_trajectory_divergence,lemma:mode_probabilities},
  there is a coupling such that
  \neqn{
    \mathbb P(E_T^c)\le K_M\delta,
    \qquad
    E_T\implies d_\Ycal(y_t,y_t^\prime)\le C_t\delta.
  }
  If also $b=b^\prime=0$, then
  $y,y^\prime\in\Ycal_{\mathrm{safe}}$.
  Convexity of $\Xcal$ and $\Zcal$ then gives, for $t\ge1$,
  \neqn{
    &E_T\cap
      \set{\mathbf1_{\Ycal_{\mathrm{safe}}}(y_t)
        \ne\mathbf1_{\Ycal_{\mathrm{safe}}}(y_t^\prime)}
    \\
    &\quad\subseteq
      \set{d_\Ycal(y_t,\partial\Ycal_{\mathrm{safe}})
        \le C_t\delta}
      \cup
      \set{d_\Ycal(y_t^\prime,\partial\Ycal_{\mathrm{safe}})
        \le C_t\delta}.
  }
  Integrating the boundary-neighborhood bound of \cref{assumption:dynamics_regularity} over 
  each preceding history therefore yields
  \neqn{
    \mathbb P\!\left(
      E_T\cap
      \set{\mathbf1_{\Ycal_{\mathrm{safe}}}(y_t)
        \ne\mathbf1_{\Ycal_{\mathrm{safe}}}(y_t^\prime)}
    \right)
    \le2L_\partial C_t\delta.
  }
  The event of safety-indicator disagreement can be conservatively estimated as
  \neqn{
    \set{b_T\ne b_T^\prime}
    \subseteq E_T^c\cup
      \bigcup_{t=1}^{T}
      \left(
        E_T\cap
        \set{\mathbf1_{\Ycal_{\mathrm{safe}}}(y_t)
          \ne\mathbf1_{\Ycal_{\mathrm{safe}}}(y_t^\prime)}
      \right),
  }
  We can then get an upper bound on the probability
  \neqn{
    \mathbb P(b_T\ne b_T^\prime)
    \le
    \left(
      K_M+2L_\partial
      \sum_{t=1}^{\bar K\tau_{\max}}C_t
    \right)\delta.
  }
  For cost, retain
  \neqn{
    \Ccal:=
      \sum_{t=0}^{T-1}\ell(x_t,u_t)
      +\ell_T(x_T)-\ell_T(x_0),
  }
  with $\Ccal^\prime$ defined analogously.
  On $E_T$,
  \neqn{
    |\Ccal-\Ccal^\prime|
    \le
    \left[
      L_\ell\sum_{t=0}^{T-1}((1+L_\pi)C_t+1)
      +L_{\ell_T}(C_T+1)
    \right]\delta.
  }
  With
  \neqn{
    \bar C:=
      \bar K\tau_{\max}
      \sup_{(x,u)\in\Xcal\times\Ucal}\ell(x,u)
      +\sup_{x\in\Xcal}\ell_T(x),
  }
  In the event of mode disagreement $E_T^c$, we get the worst case bound $|\Ccal-\Ccal^\prime|\le2\bar C$.
  Hence
  \neqn{
    |\mathbb{E}[\Ccal] -\mathbb{E}[\Ccal^\prime]|
    &\le \delta \big[
      L_\ell\sum_{t=0}^{\bar K\tau_{\max}-1}
        ((1+L_\pi)C_t+1)\\
    &\quad + L_{\ell_T}\left(
        1+\max_{0\le t\le\bar K\tau_{\max}}C_t
      \right)
      +2\bar C K_M
    \big].
  }
  This same-mode cost bound is independent of $b,b^\prime$.

  The remaining cases follow from
  \neqn{
    b=b^\prime=1&\implies\text{final-risk difference}=0,\\
    b\ne b^\prime&\implies\text{final-risk difference}
      \le1\le d_b((y,b),(y^\prime,b^\prime)),\\
    m\ne m^\prime&\implies
      d_b((y,b),(y^\prime,b^\prime))\ge C_M/D_\Ycal,\\
    m=m^\prime&\implies
      \delta\le
      D_\Ycal d_b((y,b),(y^\prime,b^\prime))
      +D(\bpi,\bpi^\prime).
  }
  Letting
  \neqn{
    L_{\mathrm{risk}}
      \ge\max\set{1,D_\Ycal/C_M},
    \qquad
    L_{\mathrm{cost}}
      \ge2\bar C D_\Ycal/C_M,
  }
  then integrating against $\lambda$, using the Markov property and
  \eqref{eq:terminal_cost_update}, and then taking the infimum over
  $\lambda\in\Gamma(P_u,P_v)$ proves both inequalities.
}

This is a key result, as it will justify the use of the $W_1$ distance for pruning by allowing us to bound the maximum difference in $J$ over the set of all admissible continuations.
It also leads to the following result on the continuity of the objective.

\begin{corollary}[Lipschitz Continuity of the Objective]
\label{thm:multi_objective_continuity}
  Fix a duration sequence
  $\Tcal=(\tau_i)_{i=1}^k$ of length
  $k\in\mathbb{N}_{\geq 0}$.
  For any $\bpi,\bpi^\prime\in\Pi^k$, let
  $\sigma:=((\pi_i,\tau_i))_{i=1}^k,\; \sigma^\prime:=((\pi_i^\prime,\tau_i))_{i=1}^k$.
  Then the components of $J$ satisfy
  \neqn{
    \left|J_\mathrm{risk}(\sigma)
      -J_\mathrm{risk}(\sigma^\prime)\right|
      &\leq L_\mathrm{risk}D(\bpi,\bpi^\prime),\\
    \left|J_\mathrm{cost}(\sigma)
      -J_\mathrm{cost}(\sigma^\prime)\right|
      &\leq L_\mathrm{cost}D(\bpi,\bpi^\prime).
  }
  Hence, for fixed $\Tcal$, each component of $J(\sigma)$ is
  Lipschitz continuous in $\bpi$ with respect to $D$.
\end{corollary}
\toggleproof{
  Apply \cref{thm:wasserstein_sensitivity} with both prefixes equal to the root's empty sequence.
  For $\bpi,\bpi^\prime\in\Pi^k$, let
  $\sigma:=((\pi_j,\tau_j))_{j=1}^k$ and
  $\sigma^\prime:=((\pi_j^\prime,\tau_j))_{j=1}^k$.
  The root laws and costs coincide, so for $i\in\set{\mathrm{risk},\mathrm{cost}}$,
  \neqn{
    |J_i(\sigma)-J_i(\sigma^\prime)|\le L_iD(\bpi,\bpi^\prime).
  }
  Thus both components, and therefore the vector-valued mapping, are Lipschitz continuous.
}

This corollary provides the required regularity conditions for the sampling-based approximation to be well posed; we've shown that for any fixed duration sequence sufficiently close policy sequences produce arbitrarily close risk and cost values.
Because $\mu_\pi$ has full support over $\Pi$, and every duration has positive sampling probability, the planner can sample arbitrarily close approximations of any target policy sequence with positive probability.

\subsection{Persistence Under Sparse Pruning}
We have just shown that nearby policy sequences have nearby objective values, but it does not guarantee that we can approximate a solution under pruning.
We next show that, for every generated node, \texttt{SparseParetoUpdate} maintains a representative in its witness whose cost does not increase and whose risk and terminal-state law remain within fixed tolerances.
\begin{theorem}[Persistent Representative]\label{thm:persistent_representative}
  Let a node $u$ of depth $1\le k_u\le \bar{K}$ be generated and assigned to witness $w$ at iteration $n$.
  $\forall n'\ge n, \; \exists r\in\Rcal_w^{(n^\prime)}$ satisfying
  \eqn{\label{eq:persistent_representative}
    c_r&\le c_u,\notag\\
    q_r&\le q_u+\min\set{\delta_q,2\delta_{\mathrm{w}}},\\
    W_1(P_r,P_u)&\le2\delta_{\mathrm{w}}.\notag
  }
\end{theorem}
\toggleproof{
  Let $j:=g(q_u)$ and consider an update of witness $w$ with input set $S=\Rcal_w\cup\set{v}$.
  For every $x\in S$, following strict dominators terminates at a non-dominated $p\in S$ satisfying $c_p\le c_x$ and $q_p\le q_x$.
  The minimum-cost representative $r$ retained in $p$'s risk cell therefore satisfies
  \neqn{
    c_r\le c_x,
    \qquad
    g(q_r)=g(q_p)\le g(q_x).
  }
  This conclusion also holds when $v$ is rejected immediately, because a dominated proposal cannot dominate any member of the mutually non-dominated incumbent set.
  Applying this to $u$ at the iteration it is generated, and thereafter to any representative that has replaced it proves inductively that, after every later completed iteration, some $r\in\Rcal_w$ satisfies $c_r\le c_u$ and $g(q_r)\le j$.
  Consequently,
  \neqn{
    q_r<(j+1)\delta_q\le q_u+\delta_q.
  }
  Moreover, both $P_u$ and $P_r$ lie within $\delta_{\mathrm{w}}$ of the immutable witness center, so
  \neqn{
    W_1(P_r,P_u)\le2\delta_{\mathrm{w}}.
  }
  Since $(y,b)\mapsto b$ is $1$-Lipschitz under $d_b$, we also have
  $q_r\le q_u+W_1(P_r,P_u)\le q_u+2\delta_{\mathrm{w}}$.
  Combining the two risk bounds proves \eqref{eq:persistent_representative}.
}

Although a generated node may later be pruned, its witness always retains a representative with no greater cost and boundedly-different risk. 
Combining this result with \cref{thm:wasserstein_sensitivity} yields the following corollary: every continuation from the retained representative remains within a fixed tolerance of the same continuation from the original node.
\begin{corollary}[Persistent Continuation]\label{thm:persistent_continuation}
  Under the conditions of \cref{thm:persistent_representative}, consider any iteration $n^\prime \ge n$.
  Then $\exists r\in\Rcal_w^{(n^\prime)}$ such that, for every common admissible continuation $\gamma$,
  \eqn{\label{eq:persistent_continuation}
    J_i(\sigma_r\Vert\gamma)
      \le J_i(\sigma_u\Vert\gamma)+2L_i\delta_{\mathrm{w}},
  }
  simultaneously for $i\in\set{\mathrm{risk},\mathrm{cost}}$.
\end{corollary}
\toggleproof{
  Select $r$ from \cref{thm:persistent_representative} and apply \cref{thm:wasserstein_sensitivity} with identical continuations and $W_1(P_r,P_u)\le2\delta_{\mathrm{w}}$.
  This gives the risk bound directly.
  For cost it gives
  \neqn{
    J_{\mathrm{cost}}(\sigma_r\Vert\gamma)
      &\le J_{\mathrm{cost}}(\sigma_u\Vert\gamma)\\
      &\quad+c_r-c_u+2L_{\mathrm{cost}}\delta_{\mathrm{w}}.
  }
  Since $c_r\le c_u$, the cost bound follows.
  The representative is chosen independently of the continuation and objective component, so these inequalities hold simultaneously.
}

\subsection{Asymptotic Near-Optimality}
We are now ready to show asymptotic near-optimality.
To show active nodes have a non-zero probability of extension, we prove that the search graph remains finite, even as $N_\mathrm{iterations} \to \infty$.
\begin{lemma}[Finite Search Graph]\label{lemma:finite_active_set}
  For each fixed $\delta_{\mathrm{w}}>0, \; \exists N_{\mathrm{wit}} < \infty$ such that every depth contains at most $N_{\mathrm{wit}}$ witnesses. Furthermore, at every iteration,
  \eqn{\label{eq:active_bound}
    |\set{v\in\Vcal_{\mathrm{active}}\mid\! k_v\!<\! \bar{K}}| \le 1 \!+\!(\bar{K}\!-\!1)N_{\mathrm{wit}}N_q.
  }
  Moreover,
  \eqn{\label{eq:graph_bound}
    |\Vcal_{\mathrm{active}}\cup\Vcal_{\mathrm{inactive}}|
      \le1+\bar{K}^2N_{\mathrm{wit}}N_q.
  }
\end{lemma}
\toggleproof{
  Since $\Ycal_b$ is compact, $\Pcal(\Ycal_b)$ is compact, hence totally
  bounded, under $W_1$.
  Let
  $\{B_{W_1}(Q_j,\delta_{\mathrm{w}}/3)\}_{j=1}^{N_{\mathrm{wit}}}$
  be a finite cover of $\Pcal(\Ycal_b)$.
  If two distinct witnesses $w,w'$ at the same depth had centers in the
  same covering ball, then
  \neqn{
    W_1(P_w,P_{w'})
    \le W_1(P_w,Q_j)+W_1(Q_j,P_{w'})
    < \delta_{\mathrm{w}},
  }
  contradicting the witness-insertion rule.
  Thus every depth contains at most $N_{\mathrm{wit}}$ witnesses.

  For $k\ge1$, let
  $\Acal_k:=\set{v\in\Vcal_{\mathrm{active}}\mid k_v=k}$.
  Since the active nodes are precisely the witness representatives and
  each witness retains at most $N_q$ representatives,
  \neqn{
    |\Acal_k|\le N_{\mathrm{wit}}N_q.
  }
  Because the root is the unique active node at depth zero,
  \neqn{
    \left|\set{v\in\Vcal_{\mathrm{active}}\mid k_v<\bar K}\right|
    \!=\!1\!+\!\!\sum_{k=1}^{\bar K-1}|\Acal_k|
    \!\le\! 1\!+\!(\bar K\!-\!1)N_{\mathrm{wit}}N_q.
  }
  Let $\Acal:=\Vcal_{\mathrm{active}}\setminus\set{v_0}$.
  Then $|\Acal|\le\bar K N_{\mathrm{wit}}N_q$.
  By the pruning rule, every surviving inactive node lies on the
  root path of some $a\in\Acal$.
  Each such path contains at most $\bar K$ nonroot nodes; hence
  \neqn{
    |\Vcal_{\mathrm{active}}\cup\Vcal_{\mathrm{inactive}}|
    \le1+\bar K|\Acal|
    \le1+\bar K^2N_{\mathrm{wit}}N_q.
    &&\qedhere
  }
}

Let $B:=1 + (\bar{K} - 1)N_\mathrm{wit}N_q$. For a target policy $\pi$, tolerance $h>0$, and duration $\tau$, the conditional probability of selecting any specified current expandable representative and sampling that duration and a policy within $h$ of $\pi$ is at least
\eqn{\label{eq:extension_probability}
  \frac{\mu_\pi(\set{\pi'\in\Pi\mid d_\Pi(\pi',\pi)<h})}
       {B\tau_{\max}}>0.
}

Having shown that the number of active nodes eligible for extension is bounded, we obtain a positive lower bound on the probability of selecting any particular node, sampling a prescribed duration, and sampling a policy within a given distance of a target policy.
We are thus ready to present the main result.

\begin{theorem}[Almost-Sure Asymptotic Near-Optimality]
\label{thm:asymptotic_near_optimality}
  With probability one, for every $\eta>0$ there exists an iteration $n_\eta<\infty$ such that, for every $N\ge n_\eta$ and every $\sigma^\ast\in\Sigma^\ast$, there exists $r\in\Vcal_{\mathrm{active}}^{(N)}$ satisfying
  \eqn{\label{eq:uniform_coverage}
    q_r
    &\!\le\!
    J_{\mathrm{risk}}(\sigma^\ast)
    \!+\! 2(\bar{K} \!-\! 1)L_{\mathrm{risk}}\delta_{\mathrm{w}}
    \!+\! \min\set{\delta_q,2\delta_{\mathrm{w}}}
    \!+\eta,\\
    c_r
    &\!\le\!
    J_{\mathrm{cost}}(\sigma^\ast)
    +2(\bar{K}-1)L_{\mathrm{cost}}\delta_{\mathrm{w}}
    +\eta.
  }
\end{theorem}
\toggleproof{
  Fix $\eta > 0$. Since the policy space $\Pi$ is compact, and $\tau_\mathrm{max}$, and  $\bar{K}$ are finite, 
  and $J$ is continuous on continuous on every fixed-duration component of $\Sigma$, $J(\Sigma)$ is compact, and thus $\Sigma^\ast\ne\emptyset$ and $J(\Sigma^\ast)$ are bounded.

  Partition a rectangle containing $J(\Sigma^\ast)$ into finitely many
  cells of width at most $\eta/2$.
  From every cell select one
  $\sigma_s^\ast\in\Sigma^\ast$, $s=1,\ldots,M$.
  Choose $h>0$ such that
  \eqn{\label{eq:near_optimality_policy_radius}
    \bar K L_i h\le\frac{\eta}{2},
    \qquad i\in\set{\mathrm{risk},\mathrm{cost}}.
  }

  Write
  \neqn{
    \sigma_s^\ast
    =
    \bigl((\pi_{s,j},\tau_{s,j})\bigr)_{j=1}^{m_s},
    \qquad 0\le m_s\le\bar K,
  }
  and let $\gamma_{s,j}$ be its suffix after segment $j$.
  For $m_s\ge1$, define
  \eqn{
    p_s:=
    \min_{1\le j\le m_s}
    \frac{
      \mu_\pi\!\left(
        \set{\pi\in\Pi:
          d_\Pi(\pi,\pi_{s,j})<h}
      \right)}
      {B\tau_{\max}}
    >0.
  }
  Positivity follows from the full support of $\mu_\pi$ and
  \eqref{eq:extension_probability}.

  Fix $s$ with $m_s\ge1$.
  Starting with $N_{s,0}:=0$, define $N_{s,j}$ recursively as the first
  iteration after $N_{s,j-1}$ that selects the root when $j=1$, or an
  earliest-created protecting representative of $u_{s,j-1}$ when
  $j\ge2$, and samples duration $\tau_{s,j}$ and a policy within $h$
  of $\pi_{s,j}$.
  The protecting representative exists by
  \cref{thm:persistent_representative} and is expandable.
  Therefore, for every $t\ge0$,
  \eqn{\label{eq:stage_waiting_bound}
    \mathbb P\!\left(
      N_{s,j}-N_{s,j-1}>t
    \right)
    \le(1-p_s)^t.
  }
  Thus every $N_{s,j}$ is finite almost surely.
  Let $u_{s,j}$ be the child generated at iteration $N_{s,j}$.

  We claim that, simultaneously for
  $i\in\set{\mathrm{risk},\mathrm{cost}}$,
  \eqn{\label{eq:near_optimality_induction}
    J_i(\sigma_{u_{s,j}}\Vert\gamma_{s,j})
    \le
    J_i(\sigma_s^\ast)
    +2(j-1)L_i\delta_{\mathrm w}
    +jL_i h.
  }
  For $j=1$, this follows from
  \cref{thm:multi_objective_continuity}, since the sampled sequence and
  $\sigma_s^\ast$ differ by less than $h$ in $D$.
  For $j\ge2$, \cref{thm:persistent_continuation} bounds the loss from
  replacing $u_{s,j-1}$ by its protecting representative by
  $2L_i\delta_{\mathrm w}$.
  From that representative, the sampled and target continuations differ
  only in segment $j$, so \cref{thm:wasserstein_sensitivity} adds at
  most $L_i h$.
  Induction proves \eqref{eq:near_optimality_induction}.

  Because there are finitely many selected sequences and each has at
  most $\bar K$ segments, almost surely all terminal nodes
  $u_{s,m_s}$ are generated by some finite iteration $N_\eta$.
  For every $N\ge N_\eta$, \cref{thm:persistent_representative} supplies
  a representative $a_s\in\Vcal_{\mathrm{active}}^{(N)}$ satisfying
  \neqn{
    q_{a_s}
    &\!\le\!
    J_{\mathrm{risk}}(\sigma_s^\ast)
    \!+\!2(m_s\!-\!1)L_{\mathrm{risk}}\delta_{\mathrm w}
    \!+\!m_sL_{\mathrm{risk}}h
    +\min\set{\delta_q,2\delta_{\mathrm w}},\\
    c_{a_s}
    &\!\le\!
    J_{\mathrm{cost}}(\sigma_s^\ast)
    \!+\!2(m_s\!\!-1)L_{\mathrm{cost}}\delta_{\mathrm w}
    \!+\!m_sL_{\mathrm{cost}}h.
  }
  For a selected empty sequence, take $a_s:=v_0$.

  Now let $\sigma^\ast\in\Sigma^\ast$ and choose the selected
  $\sigma_s^\ast$ in the same objective-space cell.
  Then
  \neqn{
    J_i(\sigma_s^\ast)
    \le J_i(\sigma^\ast)+\frac{\eta}{2}.
  }
  Since $m_s\le\bar K$ and
  $\bar K L_i h\le\eta/2$, choosing $r:=a_s$ gives
  \eqref{eq:uniform_coverage} for every $N\ge N_\eta$.
}

Thus, we have shown that SMO-SST satisfies the conditions of asymptotic near-Pareto-optimality given in \cref{def:asymptotic_near_optimality}, and that the tolerance vector $\boldsymbol{\epsilon}$ is a function of the witness radius parameter $\delta_\mathrm{w}$ and pruning sparsity parameter $\delta_q$, given explicitly by
\eqn{
  \epsilon_q &= 2(\bar{K} - 1) L_\mathrm{risk} \delta_\mathrm{w} + \min\set{\delta_q, 2\delta_\mathrm{w}}\\
  \epsilon_c &= 2(\bar{K} - 1) L_\mathrm{cost} \delta_\mathrm{w}.
} 


Thus, at fixed $\delta_{\mathrm{w}}$ and $\delta_q$, SMO-SST is asymptotically near Pareto-optimal with respect to the risk-cost Pareto front.

\section{Experimental Evaluation}\label{sec:results}
We evaluate SMO-SST in a stochastic pursuit-evasion environment with adversaries motivated by~\cite{vonmollBasicEngagementZones2024}.
For the \texttt{Propagate} step, we propagate $N_\mathrm{p}$ particles simultaneously to approximate $P_v$ and $J(\sigma_v)$, and compute the $W_1$ distance between the resulting empirical distributions.
\subsection{Environment and Dynamics}
With sampling period $\Delta t$, the ego agent is a discrete-time unicycle with state $x_t = [p_{x,t}, p_{y,t}, \theta_t, v_t]^\top$ and control $u_t = [a_t, \omega_t]^\top$,
Its dynamics are:
\neqn{
  p_{x,t+1}
    &= p_{x,t} + v_t \cos(\theta_t) \Delta t + w_{p_x,t},\\
  p_{y,t+1}
    &= p_{y,t} + v_t \sin(\theta_t) \Delta t + w_{p_y,t},\\
  \theta_{t+1}
    &= \theta_t + \omega_t \Delta t + w_{\theta,t},\\
  v_{t+1}
  &= \operatorname{clip}
  \left(
    v_t + a_t \Delta t + w_{v,t},
    -v_{\max},
    v_{\max}
  \right),
}
where $|a_t| \leq a_{\max}$ and $|\omega_t| \leq \omega_{\max}$,
subject to additive process noise
$w_{j,t} = \bar w_j (2 B_{j,t} - 1), \; B_{j,t} \sim \operatorname{Beta}(3, 3), \; j=1,\ldots,4,$
with $\bar w = [0.06, 0.06, 0.03, 0.06 v_{\max}]^\top$.
The environment contains $N$ stochastic hybrid adversaries, each with mode $m_{i,t} \in \mathbb{M}_\mathrm{adv} := \set{\textsc{Idle}, \textsc{Active}, \textsc{Terminal}}$ and state $z_{i,t} = [r_{i,t}^\top,n_{i,t}]^\top$, where $r_{i,t}\in\R^2$ and $n_{i,t} \in \mathbb{N}_{\ge0}$; the joint environment mode is $m_t=(m_{1,t},\ldots,m_{N,t})\in\mathbb{M}:=\mathbb{M}_\mathrm{adv}^N$.

When the ego is at distance $d$ from an $\textsc{Idle}$ adversary, it transitions to $\textsc{Active}$, with probability 
\neqn{
  p_\mathrm{fire}(d) = p_{\max}
  \operatorname{clip}
  \left(
    \frac{r_{\max}^2 - d^2}{r_{\max}^2 - r_{\min}^2},
    0,
    1
  \right),
}
where $p_{\max} = 0.3$, $r_{\max} = 8$, and $r_{\min} = 3$.

An \textsc{Active} adversary pursues the ego agent at constant speed $v_{\mathrm{adv},\max} = 1.5$ with simple motion, perturbed by additive noise $0.01(2\nu_{i,t}-\mathbf{1})$, where the components of $\nu_{i,t}$ are IID $\operatorname{Beta}(3, 3)$ random variables.
The state $n_{i,t}$ represents the count of pursuit steps by the projectile and is initialized at zero.
The projectile transitions to \textsc{Terminal} from \textsc{Active} at each step with probability
\neqn{
  p_\mathrm{term}(n_{i,t})
  =
  \operatorname{clip}
  \left(
    \frac{n_{i,t} - 0.8(L-1)}{0.2(L-1)},
    0,
    1
  \right)
}
where $L = 54$.
Capture occurs when the adversary is in the $\textsc{Active}$ state and within $r_\mathrm{cap} = 0.25$ of the ego.
$\textsc{Idle}$ and $\textsc{Terminal}$ adversaries remain stationary.

The space of feedback policies is parameterized by a reference waypoint sampled uniformly in the spatial domain.
The controller uses smooth saturated feedback to turn toward the waypoint and accelerate toward $v_{\max}$, with heading and speed gains of $3$ and $2$, respectively.
For the reactive policy, the desired heading blends the waypoint direction with a distance-weighted direction away from all \textsc{Active} projectiles within $R_\mathrm{rep}=8$; this influence increases as a projectile approaches the capture radius.
We smooth all normalized directions with $\epsilon_\mathrm{sm}=0.25$ and set the repulsion gain to $\kappa_\mathrm{rep}=2$; the target-only policy uses $\kappa_\mathrm{rep}=0$.

Let $z_i$ be the position of active projectile $i$, and define
\neqn{
  e_q(q,p) &:= \frac{q - p}{\sqrt{\lVert q - p \rVert_2^2 + \epsilon_\mathrm{sm}^2}},\\
  e_i(p,z_i) &:= \frac{p - z_i}{\sqrt{\lVert p - z_i \rVert_2^2 + \epsilon_\mathrm{sm}^2}},\\
  w_i(p,z_i) &:= \operatorname{clip}
    \left(
      \frac{R_\mathrm{rep} - \lVert p - z_i \rVert_2}
           {R_\mathrm{rep} - r_\mathrm{cap}},
      0,
      1
    \right),\\
  e_r(y) &:= \frac{\sum_{i:m_{i}=\textsc{Active}} w_i(p,z_i) e_i(p,z_i)} {1 + \sum_{i:m_{i}=\textsc{Active}} w_i(p,z_i)}.
}
For $\kappa_\mathrm{rep} \geq 0$, the desired direction is
\neqn{
  d_q(y)
  :=
  \frac{e_q(q,p) + \kappa_\mathrm{rep} e_r(y)}
       {\sqrt{\lVert e_q(q,p) + \kappa_\mathrm{rep} e_r(y) \rVert_2^2 + \epsilon_\mathrm{sm}^2}}.
}
Let $h_\perp(\theta) := [-\sin(\theta), \cos(\theta)]^\top$.
The feedback policy is
\eqn{
  \omega
  &=
  \omega_{\max}
  \tanh
  \left(
    3 h_\perp(\theta)^\top d_q(y)
  \right),\\
  a
  &=
  a_{\max}
  \tanh
  \left(
    2 \frac{v_{\max} - v}{v_{\max}}
  \right).
}

Execution cost is elapsed time plus a quadratic control-effort penalty with weight $0.05$, using acceleration and turn rate normalized by their respective limits.
The terminal cost increases linearly with the ego's final distance from $p_\mathrm{ref}=[60,60]^\top$ after normalization by the workspace diagonal $64\sqrt{2}$, saturating at the maximum-horizon duration $T_{\max}=\bar{K}\tau_{\max}\Delta t$.
\neqn{
  \ell(x_t,u_t)
  &:=
  \Delta t
  \left(
    1
    +
    0.05
    \left[
      \left(\frac{a_t}{a_{\max}}\right)^2
      +
      \left(\frac{\omega_t}{\omega_{\max}}\right)^2
    \right]
  \right),\\
  \ell_T(x_T)
  &:=
  T_{\max}
  \operatorname{clip}
  \left(
    \frac{\lVert p_T - p_\mathrm{ref} \rVert_2}{64 \sqrt{2}},
    0,
    1
  \right),
}

We consider a $64 \times 64$ reference region containing $N = 8$ adversaries.
The ego state is initialized at $\bmat{0, 0, 0, 0}^\top$.

\subsection{Evaluation \& Results}
\begin{table*}[!t]
  \centering
  \caption{Equal wall-clock-matched results over $N=30$ paired environments. Reported statistics are means, with $95\%$ CI.}
  \begin{tabular}{lrrrr}
    \toprule
    Method & nHV (CI) $\uparrow$ & Fresh nHV (CI) $\uparrow$ & Iterations & Nodes \\
    \midrule
    Reactive SMO-SST & 0.759 [0.739, 0.769] & 0.757 [0.735, 0.767] & 336405 & 966 \\
    Target-only SMO-SST & 0.757 [0.734, 0.768] & 0.754 [0.731, 0.765] & 334482 & 1002 \\
    Reactive SMO-RRT & 0.714 [0.707, 0.719] & 0.713 [0.706, 0.719] & 529380 & 529381 \\
    Target-only SMO-RRT & 0.709 [0.701, 0.715] & 0.708 [0.700, 0.714] & 518807 & 518808 \\
    \bottomrule
  \end{tabular}
  \label{tab:results_table}
\end{table*}
\begin{figure*}[h]
  \centering
  \includegraphics[width=0.63\textwidth]{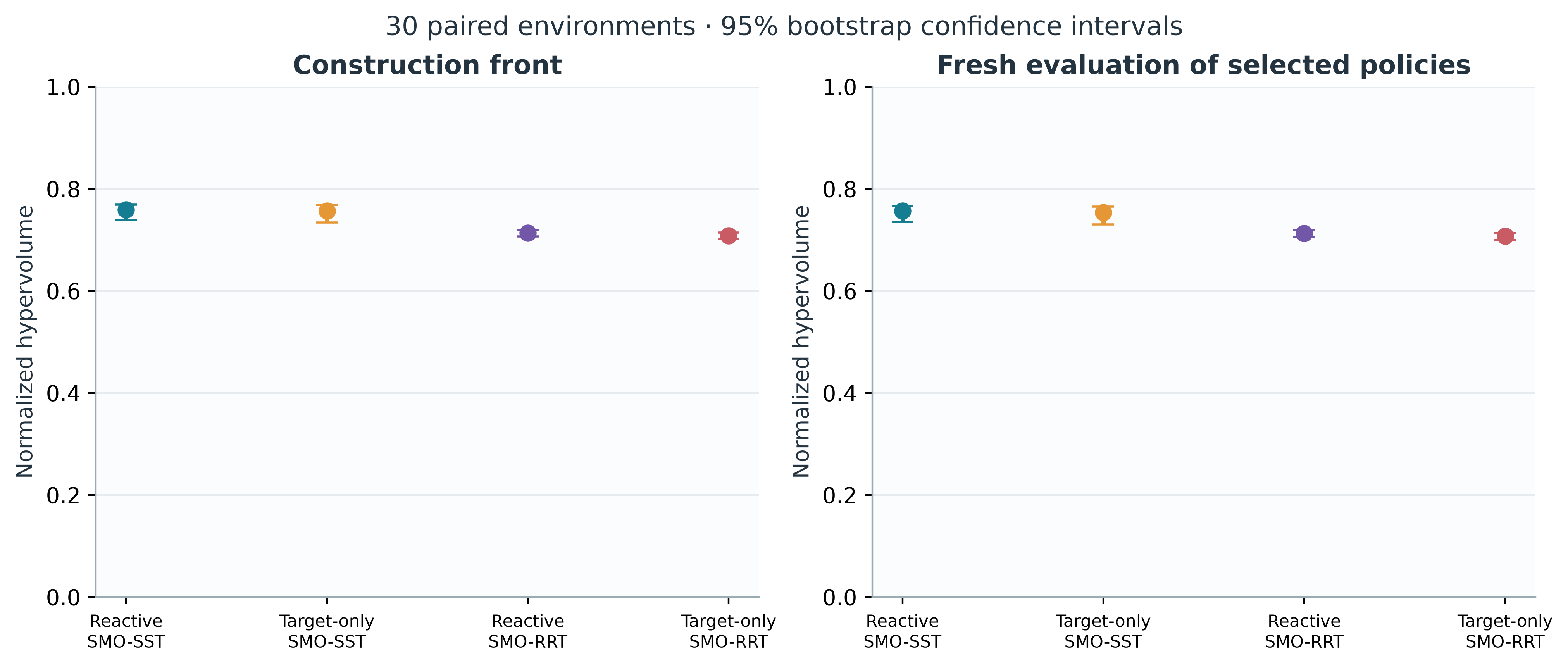}
  \includegraphics[width=0.36\textwidth]{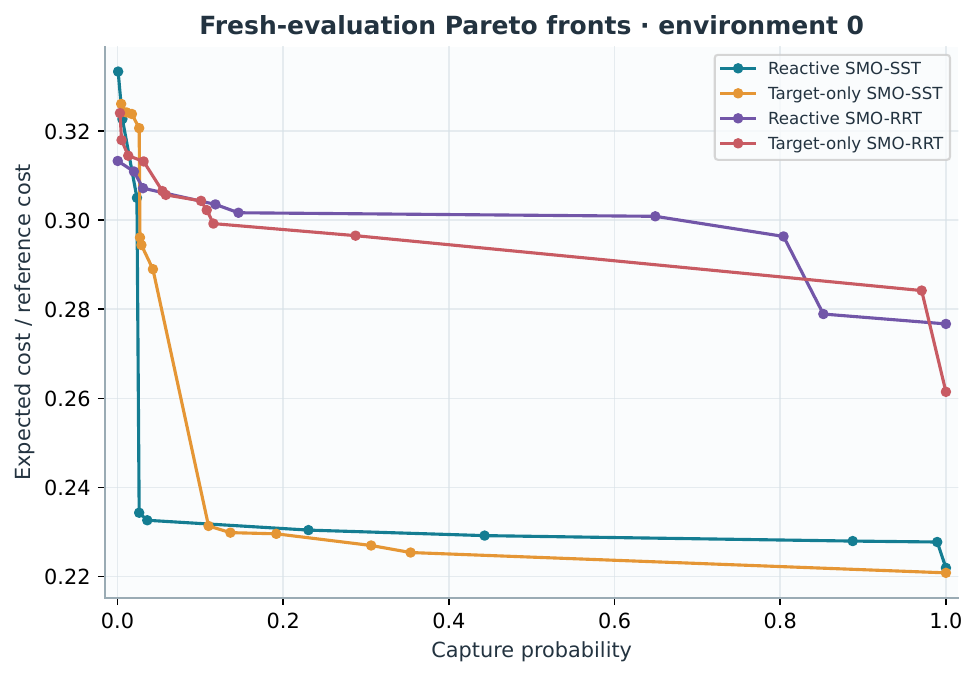}
  \caption{(Right) Shows the mean \& 95\% Bootstrap CIs for normalized hypervolume (nHV) for each algorithm over 30 paired environments. (Middle) the nHV of the solutions after being re-evaluated with more particles. (Right) Example Pareto-front of solutions in one environment broken down by algorithm. We see the SST variants recover solutions of significantly lower cost at the same risk levels. }
  \label{fig:pareto_front}
\end{figure*}

Simulations are executed on a MacBook Pro with an M5-Pro and 48GB RAM.
Because the environment violates assumptions of prior methods, we use four internal ablations to isolate the effects of pruning and reactive feedback:
\begin{enumerate}
  \item \emph{Reactive SMO-SST} uses the planner with $\kappa_\mathrm{rep} = 2$.
  \item \emph{Target-only SMO-SST} uses the same planner and policy parameterization with $\kappa_\mathrm{rep} = 0$.
  \item \emph{Reactive SMO-RRT} disables witness assignment and pruning, but retains the policy parameterization.
  \item \emph{Target-only SMO-RRT} uses $\kappa_\mathrm{rep} = 0$.
\end{enumerate}
The target-only comparisons isolate the effect of reactive feedback, while the SMO-RRT comparisons isolate the computational and solution-quality effects of pruning.
SST uses $\delta_\mathrm{w}=0.25$ and $\delta_\mathrm{q}=0.01$.

For the returned set of non-dominated policies, normalized hypervolume is computed using the fixed reference point $(1, \bar{C})$ and an ideal point $(0, 0)$.

The main study uses $N=30$ randomly generated environments and $N_\mathrm{p} = 256$ particles.
We perform a wall-clock-matched comparison, with each planner given $2$ minutes.
\Cref{tab:results_table} reports normalized hypervolume (nHV) of the returned solutions as evalauted during runtime, the Fresh nHV is the result of re-evaluating the returned policies using $N=4096$ particles, iteration count (mean), and retained node count (mean), showing that SMO-SST outperforms the non-pruning baselines while retaining $99.8\%$ fewer nodes.
The results are additionally visualized in \cref{fig:pareto_front}.

\subsection{Effects of Sparsity Parameters on Performance}
\begin{figure*}[ht]
  \centering
  \includegraphics[width=1.0\textwidth]{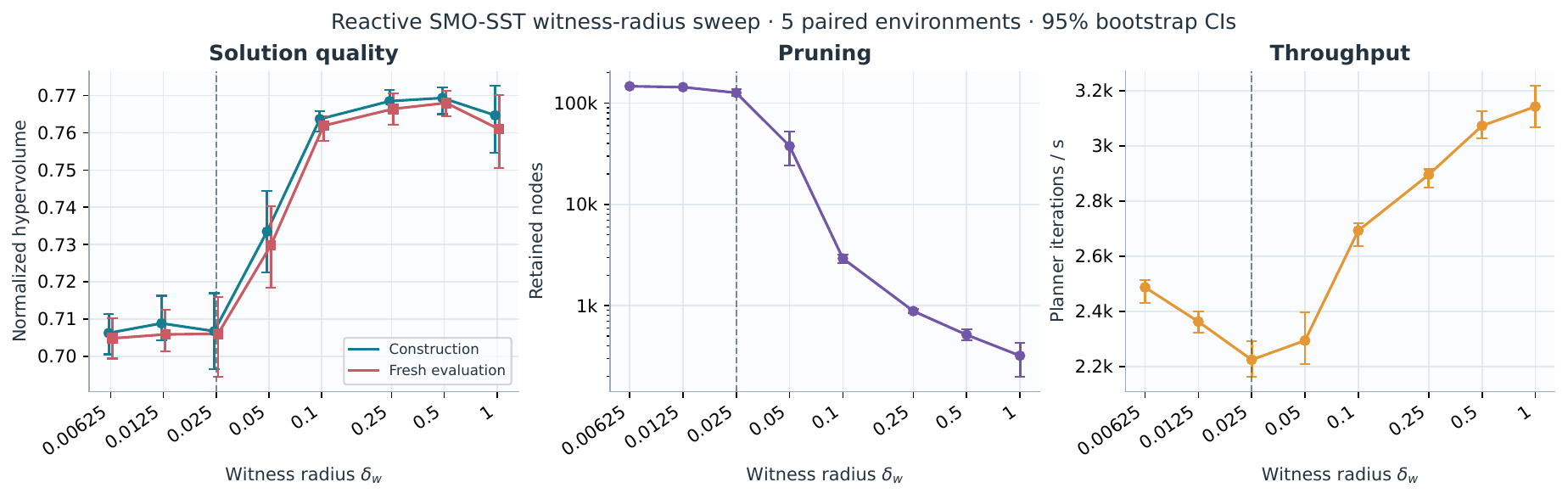}
  \caption{The results of a parameter sweep over $\delta_w$. We see that at $\delta_w \le 0.025$, very little pruning occurs. The amount of pruning increases isgnificantly between $0.025 < \delta_w \le 0.1$, and then begins to plateau again for $0.1 < \delta_w \le 1.0$. Solution quality also begins to decline with $\delta_w=1$.} 
  \label{fig:sparsity_performance}
\end{figure*}

\begin{figure*}[t]
  \centering
  \includegraphics[width=1.0\textwidth]{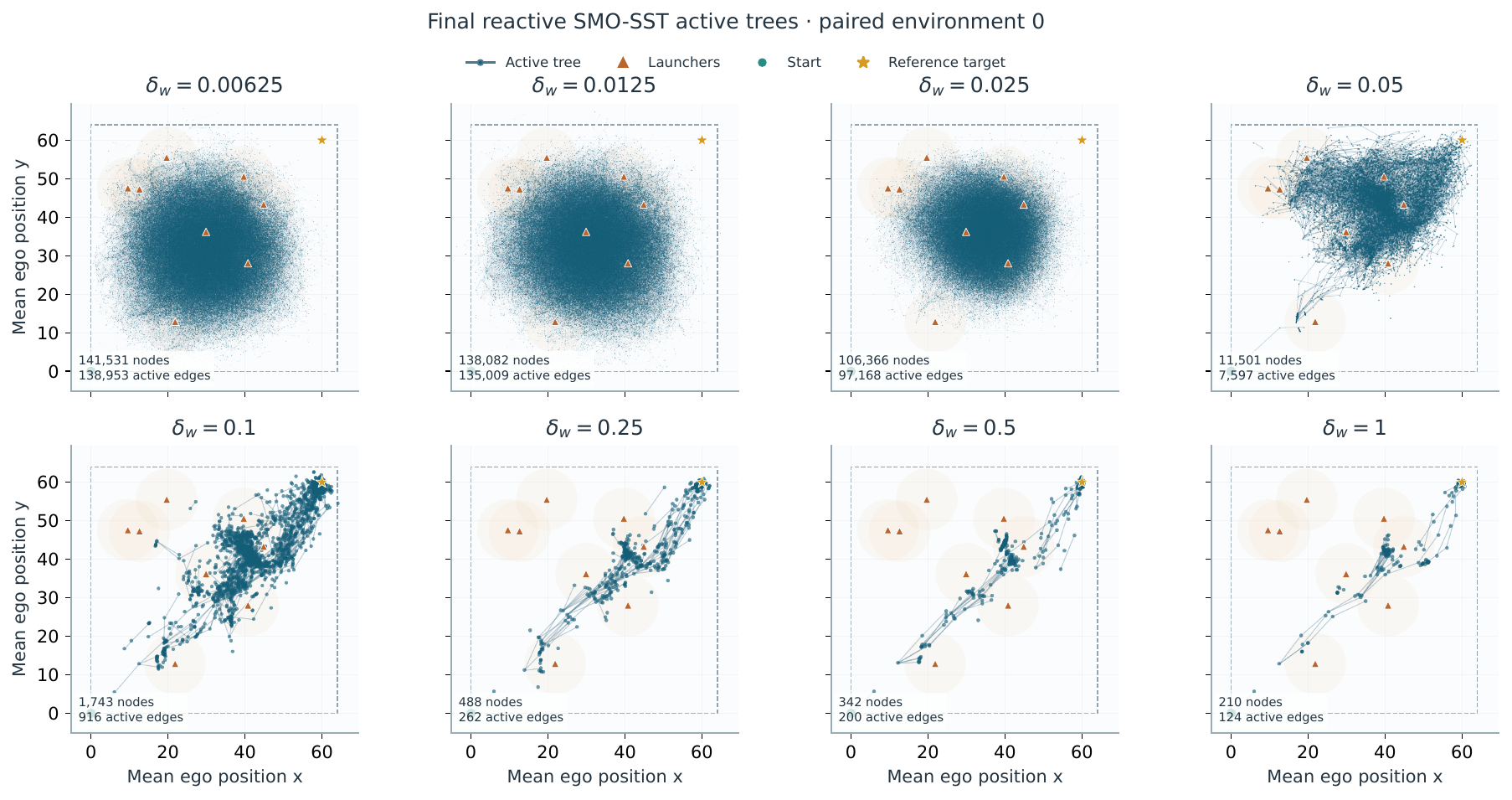}
  \caption{Figure shows the trees produced by SMO-SST in the same environment at different sparsity parameters $\delta_w$. The retained node count drops significantly, and the tree is heavily biased towards the Pareto front of solutions.}
  \label{fig:delta_w_trees}
\end{figure*}

We perform additional analysis on the effects of the sparsity parameters $\delta_\mathrm{w}$ and $\delta_\mathrm{q}$ on the performance of SMO-SST in the same pursuit-evasion environment.
We first evaluate $\delta_\mathrm{w} = \bmat{0.00625, 0.0125, 0.025, 0.05, 0.1, 0.25, 0.5, 1.0}^\top$
in $N=5$ trials in heldout matched environments.
Results are reported in \cref{fig:sparsity_performance}.
We see that very little pruning occurs for $\delta_w \le 0.025$, and over $100k$ nodes (almost all generated) are retained.
For $0.025 \le \delta_w \le 0.1$, we see a significant increase in the effects of pruning, with retained node count being reduced from over $100k$ down to $~4k$. 
For $\delta_w \ge 0.1$, solution quality continues to increase slightly, and we see a significant increase in throughput, likely due to nodecount dropping below $1k$. This coincides with an increase in throughput, which increases as $\delta_w$ increases from a minimum at $\delta_w = 0.025$.
\Cref{fig:delta_w_trees} shows the resulting tree structure under different values $\delta_w$. The tree becomes very biased towards the goal state, corresponding to a bias towards low-cost solutions.


\section{Conclusion}\label{sec:conclusion}
  We presented SMO-SST, a stochastic multi-objective kinodynamic planner for environments modeled by reactive stochastic hybrid systems.
  Prescribing a chance-constraint threshold can conceal useful safety-performance tradeoffs, so we jointly minimize violation probability and expected execution cost over finite feedback-policy sequences.
  We show the proposed planner is asymptotically near-Pareto-optimal, and provide an explicit approximation bound.
  Across paired environments, SMO-SSt achieved higher nHV that its corresponding non-pruning SMO-RRT baselines while retaining approximately $99.8\%$ fewer nodes.
  Future work will investigate asymptotic optimality, and guarantees for the particle implementation.


\bibliographystyle{IEEEtran}
\bibliography{biblio, 109_pareto_rrt}



\end{document}